\documentclass{article}
\usepackage{spconf,amsmath,graphicx}
\usepackage{dsfont}
\usepackage{xcolor}
\usepackage{caption}
\usepackage{booktabs}
\usepackage{multirow}
\usepackage{bm}
\usepackage{arydshln}
\usepackage{subcaption}
\usepackage{flushend}

\title{Trainable Time Warping: \\aligning time-series in the continuous-time domain}
\name{Soheil Khorram, Melvin G McInnis, Emily Mower Provost}
\address{University of Michigan, Ann Arbor, USA}
\begin{document}
\ninept
\maketitle

\begin{abstract} 
DTW calculates the similarity or alignment between two signals, subject to temporal warping. However, its computational complexity grows exponentially with the number of time-series. 
Although there have been algorithms developed that are linear in the number of time-series, they are generally quadratic in time-series length.  The exception is generalized time warping (GTW), which has linear computational cost.  Yet, it can only identify simple time warping functions.  There is a need for a new fast, high-quality multisequence alignment algorithm. We introduce \textit{trainable time warping (TTW)}, whose complexity is linear in both the number and the length of time-series. TTW performs alignment in the continuous-time domain using a sinc convolutional kernel and a gradient-based optimization technique. We compare TTW and GTW on $85$ UCR datasets in time-series averaging and classification. TTW outperforms GTW on $67.1\%$ of the datasets for the averaging tasks, and $61.2\%$ of the datasets for the classification tasks.
\vspace{-3pt} 

\end{abstract}
%

\vspace{-2pt}
\section{Introduction}
\label{sec:intro}
\vspace{-1pt}

Time-series alignment is an important question in many different applications, including bioinformatics~\cite{aach2001aligning}, computer vision~\cite{rath2003word}, speech recognition~\cite{muda2010voice}, speech synthesis~\cite{khorram2013speech, khorram2014context} and human action recognition~\cite{sempena2011human}. The most commonly used method to perform this alignment is dynamic time warping (DTW), which uses the Bellman's dynamic programming technique to search for the optimal time warping~\cite{morel2018time, schultz2018nonsmooth}. However, it is difficult to apply DTW over large sets of time-series because the computational complexity\footnote{Computational complexity refers to both time and space complexities throughout this paper.} of DTW grows exponentially with the number of time-series to be aligned~\cite{jain2018asymmetric, brill2018exact}. 

In this paper, we introduce a new time warping algorithm, called \textit{trainable time warping (TTW)} with linear computational complexity in both $N$ and $T$, where $N$ and $T$ are the number and the length of time-series, respectively\footnote{For the sake of simplicity in notations, we assume time-series have the same length. Extending TTW to variable-length signals is straightforward. One method is to interpolate the signals to the maximum length of them.}. TTW first aligns the input sequences and then takes an average over the synchronized sequences to calculate the centroid signal. We propose a sinc convolutional kernel that is able to warp signals in a nonlinear (elastic) manner. TTW uses this convolutional kernel along with a gradient-based optimization technique to synchronize multiple time-series.

DTW extensions for multiple time-series include a progressive DTW-based method, called NonLinear Alignment and Averaging Filters (NLAAF)~\cite{gupta1996nonlinear}. In each iteration of NLAAF, time-series are randomly grouped into a set of pairs and then each pair is fused using the DTW averaging. This procedure is repeated until the centroid time-series is achieved.  However, although NLAAF is practical for averaging multiple sequences, it has two problems: (1) ordering - the centroid is sensitive to the order of pairing time-series and (2) error propagation - there is no feedback in the algorithm to reduce the errors that occur at early steps. 

Many papers presented methods to address these two weaknesses. Niennattrakul et al.~\cite{niennattrakul2009shape}, addressed the first problem, introducing a method that uses hierarchical clustering to automatically create a reasonable order for pairing time-series. Petitjean et al.~\cite{petitjean2011global, soheily2015progressive}, addressed both, introducing an averaging method, DTW Barycenter Averaging (DBA), which iterates over two steps: (1) aligning: DBA computes a DTW mapping between the current centroid and each input time-series and (2) updating: DBA uses the mappings obtained from the previous step to align the time-series and update the centroid sequence. However, DBA is sensitive to the quality of the initial average signal. When the initial average is considerably different from the best solution, DBA is highly likely to converge to a weak local minimum~\cite{cuturi2017soft}. 
Extensions have been proposed to improve the performance of the DBA algorithm (e.g., soft-DTW~\cite{cuturi2017soft}, stochastic subgradient (SSG)~\cite{schultz2018nonsmooth} and constraint DTW~\cite{morel2018time}). However, all have the computational complexity of $\mathcal{O}(NT^2)$. The complexity is quadratic in $T$, which is problematic in many applications that must process long-term time-series~\cite{salvador2007toward}.





To the best of our knowledge, generalized time warping (GTW) \cite{zhou2012generalized, zhou2009canonical} is the only algorithm that is able to align multiple signals with linear complexity in $T$. GTW approximates the optimal temporal warping by linearly combining a fixed set of monotonic basis functions. They introduced a Gauss-Newton-based procedure to learn the weights of the basis functions. However, in the cases where the temporal relationship between the time-series is complex, GTW requires a large number of complex basis functions to be effective; defining these basis functions is very difficult. 

We evaluate the efficacy of TTW on $85$ datasets of the UCR archive~\cite{UCRArchive} for two applications: DTW averaging and time-series classification. We implement GTW as the baseline system, because GTW has similar computational complexity. The results show that TTW is able to outperform GTW on most of the datasets. 

\vspace{-6pt}
\section{Problem Setup}
\vspace{-3pt}
\label{sec:ProblemSetup}

Given a set of $N$ time-series of length $T$, $\bm{X}=\{\bm{x}_1,\scalebox{0.75}{\dots},\bm{x}_{N}\}$ where $\bm{x}_n=\bm{[}\ x_n\scalebox{0.9}{[1]},\scalebox{0.75}{\dots},x_n\scalebox{0.9}{[T]}\ \bm{]}$, the goal of DTW averaging is to estimate an optimal average signal $\bm{\hat y}$ that minimizes the following cost:
\\\vspace{-5pt}
\[
\bm{\hat y} = \arg\min_{\bm{y}} \sum_{n=1}^{N} \mathcal{D}_{dtw}(\bm{x}_n, \bm{y}), \tag{1}
\]\vspace{-5pt}
\\
where $\mathcal{D}_{dtw}(\bm{x}_n, \bm{y})$ is the DTW distance between $\bm{x}_n$ and $\bm{y}$~\cite{petitjean2011global, keogh2005exact, jain2017optimal}. In this paper, we approximate $\bm{\hat y}$ by estimating a set of synchronized signals, $\bm{\widetilde X}=\{\bm{\widetilde x}_1,\scalebox{0.75}{\dots},\bm{\widetilde x}_{N}\}$, where $\bm{\widetilde x}_n=\bm{[} \widetilde{x}_n\scalebox{0.9}{[1]},\scalebox{0.75}{\dots},\widetilde{x}_n\scalebox{0.9}{[T]}\ \bm{]}$, and then taking average of this set: $\bm{\hat y}= \frac{1}{N}\sum_{n=1}^{N}\bm{\widetilde x}_n$.



\vspace{-6pt}
\section{Methods}\label{sec:methods}
\vspace{-3pt}

In this section, we first describe how the shifted sinc kernel can be used to introduce a learnable uniform shift to a time-series. We then extend to a more general non-linear time warping case. Finally, we introduce the TTW algorithm.

\vspace{-6pt}
\subsection{Shifted Sinc Kernel}\label{sec:Shifted Sinc Kernel}
\vspace{-3pt}

We first consider a simple case where the time alignment between the signals can be performed by applying a uniform delay $\tau_n$ (i.e., $\widetilde{x}_n[t]=x_n[t-\tau_n]$). We can implement this alignment by convolving the input signal $\bm{x_n}$ with a shifted dirac-delta, $\delta[t-\tau_n]$, kernel:
\vspace{-4pt}
\[
\forall n\in\{1,\scalebox{0.75}{\dots},N\},\ \ \ 
\widetilde{x}_n[t] = x_n[t] * \delta[t-\tau_n], \tag{2}
\]
\vspace{-1pt}
where $*$ is the convolution operation.

In order to perform temporal alignment, we must estimate the best value for the shift parameter $\tau_n$, which can be done through a search algorithm~\cite{mariooryad2015correcting}. However, search algorithms are inefficient when there are multiple input signals, because the search space grows exponentially with the number of the signals (similar to the standard DTW). Gradient-based optimization techniques are commonly used to deal with very large search spaces. However, $\tau_n$ is not learnable in this way because: (1) $\tau_n$ is an integer, while gradient-based optimization strategies have been designed for a continuous search space and (2) $\delta(t)$ is not differentiable whenever $t=0$.

As a solution, we propose to transfer the time-series to the continuous-time domain and perform the temporal alignment in that domain. In this approach, $\tau_n$ is applied to a continuous time-series and can be any real number. To implement this idea, three main steps are needed: (1) interpolation: transferring the input signals from discrete to continuous-time domains; (2) shifting: shifting the signals in the continuous-time domain; (3) sampling: returning the signals to the original discrete-time domain. 

\vspace{2pt}
\textbf{Interpolation} -- In the first step, we use sinc interpolation to transfer to the continuous-time domain. Ideal sinc interpolation preserves all frequency components of the input signals~\cite{yaroslavsky1997efficient}. The relationship between discrete and continuous time signals in the sinc interpolation technique is\footnote{In this paper, we use $[.]$ for descrete-time signals and $(.)$ for continuous-time signals. For example, $f[m]$ is the $m$-th sample of a descrete-time signal $\bm{f}$, and $\bar f(t)$ is the value of a continuous-time signal, $\bm{\bar f}$, at time $t$.}:
\\[-7pt]
\[\bar x_n(t)=\sum_{m=1}^{T}x_n[m]sinc
\begin{pmatrix}
\frac{t}{T_s}-m
\end{pmatrix}
, \tag{3}\]
\\[-7pt]
where $x_n[m]$ is the $m$-th sample of the $n$-th signal, $\bar x_n(t)$ is an interpolation of the $n$-th signal at time $t$, $T_s$ is the sampling rate and the $sinc$ function is defined as:
\\[-5pt]
\[
sinc(t) =
\left\{\begin{matrix}
\frac{sin({\pi}t)}{({\pi}t)} & t \neq 0 \\ 
1 & t = 0
\end{matrix}\right.
. \tag{4}\]

\textbf{Shifting} -- In the second step, we temporally align the continuous-time signals obtained from the previous step. In the simplest case, the alignment is performed by applying a fixed shift, i.e.,
\\[-6pt]
\[\widetilde{x}_n(t) = \bar x_n(t-{\tau_n}T_s)=\sum_{m=1}^{T}x_n[m]sinc
\begin{pmatrix}
\frac{t}{T_s}-\tau_n-m
\end{pmatrix}. \tag{5}\]
\\[-6pt]
In this equation, the size of the shift is $\tau_n$ samples or ${\tau_n}T_s$ seconds. We can simplify Equation~(5) using properties of the sinc function:
\\[-6pt]
\[
\begin{matrix}
\widetilde{x}_n(t) = \begin{pmatrix}
\sum_{m=1}^{T}x_n[m]sinc(\frac{t}{T_s}-m)
\end{pmatrix}
*
sinc
\begin{pmatrix}
\frac{t}{T_s}-\tau_n
\end{pmatrix}
\\[-6pt]\\
\widetilde{x}_n(t) = \bar x_n(t) *
sinc
\begin{pmatrix}
\frac{t}{T_s}-\tau_n
\end{pmatrix}.
\end{matrix}
 \tag{6}\]
\\[-6pt]
This equation shows that we can use a shifted low-pass (sinc) filter to introduce a fixed shift to the interpolated signals.

\vspace{2pt}
\textbf{Sampling} -- Finally, in the third step, we transfer the shifted signals to the original discrete-time domain using uniform sampling. Uniform sampling is performed by replacing $t$ with $tT_s$ in Equation~(6) which results in:
\\[-6pt]
\[\widetilde{x}_n[t] = x_n[t] * sinc(t-\tau_n) = \sum_{m=1}^{T}{x_n[m] sinc(t-\tau_n-m)}. \tag{7}\]
\\[-6pt]
This equation is the final expression for the shifted sinc kernel. In the case of integer $\tau_n$, the kernel (Equation~(7)) is equal to the dirac-delta function in the discrete-time domain (Equation~(2)), and therefore the sinc kernel just introduces a fixed shift. In other cases, the kernel interpolates the signal to continuous-time domain, introduces a shift there and returns the signal to the discrete-time domain. 

Equation~(7) shows that introducing a shift in the continuous-time domain is equivalent to convolving the signals with a shifted sinc kernel in the discrete-time domain. Therefore, \textit{to align the signals with a simple shift, we just need to convolve them with the shifted sinc kernels (Equation~(7)) and learn the shift parameters $\tau_n$}. The sinc kernel gives us the ability to introduce real-valued shifts to discrete-time signals. It is also a differentiable function and can therefore be used in gradient-based optimization techniques.

\vspace{-6pt}
\subsection{Non-linear Temporal Warping Using Sinc Kernel}
\vspace{-3pt}
\label{sec:Time-Varying Shifted Sinc Kernel}
The shifted sinc kernel is able to apply a fixed shift; however, a fixed shift is generally not sufficient for aligning time-series. In DTW-based algorithms, alignment is performed by applying an \textit{elastic (non-linear)} transform to the time axis of the signals. This transform is called time warping or time alignment function. Inspired by these algorithms, we extend our linear transform ($t\xrightarrow{}t-\tau_n$), to a more general nonlinear transform ($t\xrightarrow{}\tau_n[t]$) by replacing the term $t-\tau_n$ with $\tau_n[t]$ in Equation~(7):
\\[-6pt]
\[\widetilde{x}_n[t] = \sum_{m=1}^{T}x_n[m]sinc(\tau_n[t]-m). \tag{8}\]
\\
We represent the set of all warping functions with
$\bm{\mathcal{T}}=\{\bm{\tau}_1,\scalebox{0.75}{\dots},\bm{\tau}_N\}$,
where
$\bm{\tau}_n=[\tau_n\scalebox{0.9}{[1]},\scalebox{0.75}{\dots},\tau_n\scalebox{0.9}{[T]}]$
is the warping function that we use to align the $n$-th signal. Equation~(8) states that the $t$-th sample of the synchronized signal, $\widetilde{x}_n[t]$, will be equal to an interpolation of the original signal at time $\tau_n[t]$ (i.e., $\widetilde{x}_n[t]=\bar x_n(\tau_n[t])$). Note that we use ideal sinc interpolation to estimate the value of the original signal at time $\tau_n[t]$. 

However, calculating the synchronized signals through Equation~(8) is computational expensive. For a signal of length $T$, it needs $\mathcal{O}(T^2)$ arithmetic operations, which is problematic when the algorithm processes long-term signals. In order to overcome this problem, we approximate the ideal sinc with a windowed sinc function. We know that the sinc function is a sine wave that decays by increasing the time; more than $99\%$ of its power lies in the range of $[-10, 10]$. Therefore, we assume $sinc(t)=0$ whenever $|t|>10$. Using this assumption we can approximate Equation (8) by:
\newcommand{\floor}[1]{\lfloor #1 \rfloor}
\\[-6pt]
\[
\widetilde{x}_n[t] = 
\sum_{m=\floor{\tau_n[t]}-10}^{\floor{\tau_n[t]}+10}x_n[m]sinc(\tau_n[t]-m)
, \tag{9}\]
\\[-6pt]
where $\floor{.}$ denotes the floor operator. In other words, we use a truncated sinc (windowed sinc with a rectangular window) to reduce the computational complexity of Equation~(9) from $\mathcal{O}(T^2)$ to $\mathcal{O}(T)$. 

\subsection{DTW Averaging Using Sinc Kernel}

To accomplish DTW averaging, we first synchronize the given time-series through an optimal set of time warping functions, $\bm{\mathcal{T}_{opt}}$; we then take average of the synchronized sequences. We estimate the optimal set of time warping functions by numerically solving the following optimization problem using the Adam algorithm~\cite{kingma2014adam}: 
\\[-6pt]
\[
\bm{{\mathcal{T}_{opt}}} = \arg\min_{\bm{\mathcal{T}}} \mathcal{D}(\widetilde{\bm{X}})\ \ \ \ \ S.T.\ \ \ \ \ \textit{\textbf{\scalebox{1}{DTW\ Constraints}}},
\tag{10}
\]
\[
\mathcal{D}(\widetilde{\bm{X}})=
\frac{1}{NT}\sum_{n=1}^{N}\sum_{t=1}^{T}
(\widetilde{x}_n[t]
-y[t])^2, \ \ 
y[t]=
\frac{1}{N}\sum_{n=1}^{N}
\widetilde{x}_n[t],\]
where $\mathcal{D}(\widetilde{\bm{X}})$ is the within-group mean square error (MSE) of the synchronized signals, $\widetilde{\bm{X}}$, and $y[t]$ is the $t$-th sample of the average signal. According to this Equation, we find the optimal alignments that satisfy the DTW conditions and minimizes the within-group distance of the synchronized signals. The DTW conditions guarantee that the synchronized sequences preserve the shape of the original time-series. The next section will explain these conditions in detail.

\vspace{-6pt}
\subsection{Time Warping Constraints}
\vspace{-3pt}
\label{sec:Time Warping Constraints}

In this section, we ensure that our time warping algorithm satisfies the DTW constraints including \textit{continuity}, \textit{monotonicity} and \textit{boundary} conditions~\cite{zhou2012generalized}. This mitigates concerns regarding unreasonable choices of $\bm{\mathcal{T}}$, which could otherwise change the shape of the input time-series in an unsatisfactory manner.

\vspace{2pt}
\textbf{Continuity}: 
Sudden changes in the warping function distort the overall shape of the input time-series~\cite{zhou2012generalized, morel2018time}. In order to avoid sudden changes, we enforce a smoothness over $\tau_n[t]$. There are different methods for generating a smooth sequence. In this paper, we propose to estimate the first $K$ coefficients of the \textit{discrete sine transform (DST)} of the warping functions instead of estimating all samples~\cite{gupta1990fast}. This guarantees the smoothness of the alignment by eliminating high frequency components. In other words, we assume that the warping functions can be written as:
\vspace{-7pt}
\[
\tau_n[t]=t+\sum_{k=1}^{K}a_k^n sin\left(\frac{\pi{k}(t-1)}{(T-1)}\right),\:\: \tag{11}\vspace{-5pt}
\]
where $\bm{a}_n=[a_1^n,\scalebox{0.75}{\dots},a_K^n]$ is the DST coefficients of $\tau_n[t]$. In the alignment process, we find these coefficients instead of $\tau_n[t]$. The parameter $K$ (number of sine components) controls the smoothness of the warping function such that with a smaller value of $K$ we generate a smoother warping function. By increasing the value of $K$, we can capture higher-frequency variations of the input signals. However, our experiments show that TTW with a higher value of $K$ is more likely to converge to a weak local minimum. In addition, The DST expressed by Equation~(11) significantly reduces the number of the learning parameters, which enables us to leverage more complex (e.g., quasi-Newton) optimization algorithms. 

\vspace{2pt}
\textbf{Monotonicity}: Time warping algorithms must preserve the order of the samples. This constraint guarantees that the time warping algorithm provides consistent temporal changes between input and output sequences. The warping function must be monotonically increasing to satisfy this constraint~\cite{zhou2012generalized} (i.e., for all $t\in\{2,\scalebox{0.75}{\dots},T\}$, $\Delta\tau_n[t]\geq0$). In the TTW algorithm, we enforce this constraint in our optimization process; after each iteration (update) of the Adam algorithm, we traverse our warping functions from beginning to end; if a sample does not follow the monotonicity condition, we set the sample to its previous sample; more precisely, for all $t\in\{2,\scalebox{0.75}{\dots},T\}$ and $\Delta\tau_n[t]<0$, we set $\tau_n[t]$ to $\tau_n[t-1]$.

\vspace{2pt}
\textbf{Boundary conditions}:  $\tau_n[1]=1$ and $\tau_n[T]=T$. This constraint (along with the other constraints) guarantees that the aligned signal contains all parts of the original signal~\cite{silva2016prefix}. Our algorithm is guaranteed to satisfy this constraint because the DST transform expressed by Equation~(11) generates $1$ and $T$ for the starting ($\tau_n[1]$) and the ending ($\tau_n[T]$) points of the warping functions. 

\vspace{-6pt}
\subsection{Trainable Time Warping Algorithm}
\vspace{-3pt}
\label{sec:Trainable Time Warping}
The goal is to synchronize a set of $N$ time-series of length $T$, $\bm{X}=\{\bm{x}_1,\scalebox{0.75}{\dots},\bm{x}_{N}\}$ where $\bm{x}_n\in\mathds{R}^T$, and fuse them into a single centroid time-series $\bm{y}\in\mathds{R}^T$. To do so, we define $K$ learning parameters, $\bm{a}_n=\{a_1^n,\scalebox{0.75}{\dots},a_K^n\}$, for each input signal $\bm{x}_n$ and initialize them with zero. TTW finds the optimal set of learning parameters through iterating over the following forward, backward and update steps. 

\vspace{2pt}
\textbf{Forward}: takes the given signals $\bm{X}$ and current values of the learning parameters, $\bm{A}=\{\bm{a}_1,\scalebox{0.75}{\dots},\bm{a}_{N}\}$ as the input to compute the current average sequence. In the forward step, TTW generates $\bm{\mathcal{T}}$ using Equation~(11); fixes the monotonicity problems of $\bm{\mathcal{T}}$ using the algorithm explained in Section~\ref{sec:Time Warping Constraints} (Monotonicity); generates $\bm{\widetilde{X}}$ using Equation~(9); and calculates the within-group distance $\mathcal{D}$ and the current mean signal $\bm{y}$ using Equation~(10).

\vspace{2pt}
\textbf{Backward}: takes all intermediate signals of the forward step to generate the derivatives of the within-group distance $\mathcal{D}(\widetilde{\bm{X}})$ with respect to the learning parameters $\bm{A}$, (i.e., $\frac{\partial \mathcal{D}}{\partial \bm{A}}$). $\frac{\partial \mathcal{D}}{\partial \bm{A}}$ is an $N$-by-$K$ matrix that can be calculated through the following expression:
\vspace{-3pt}
\[
\frac{\partial \mathcal{D}}{\partial \bm{A}}_{N \times K}=
\frac{\partial \mathcal{D}}{\partial \bm{\mathcal{T}}}_{N \times T} \times
\frac{\partial \bm{\mathcal{T}}}{\partial \bm{A}}_{T \times K}, \tag{12}
\vspace{-2pt}
\]
where $\times$ denotes matrix multiplication and $\frac{\partial \mathcal{D}}{\partial \bm{\mathcal{T}}}$ is an $N$-by-$T$ matrix that contains the derivatives of the within-group distance $\mathcal{D}(\widetilde{\bm{X}})$ with respect to the warping functions; the following equation calculates the $(n,t)$-th element of this matrix:
\vspace{-3pt}
\[
\frac{\partial \mathcal{D}}{\partial \tau_n[t]}=
\frac{2}{NT}(\widetilde{x}_n[t]
-y[t])\frac{\partial \tilde x_n[t]}{\partial \tau_n[t]},
\]
\vspace{-8pt}
\[
\frac{\partial \tilde x_n[t]}{\partial \tau_n[t]}=
\sum_{m=\floor{\tau_n[t]}-10}^{\floor{\tau_n[t]}+10}x_n[m]sinc'(\tau_n[t]-m), \tag{13}
\vspace{-2pt}
\]
where $sinc'$ is the derivative of the sinc function. In addition, $\frac{\partial \bm{\mathcal{T}}}{\partial \bm{A}}$ denotes the derivatives of the warping functions with respect to the learning parameters which is a $T$-by-$K$ matrix with the following $(t,k)$-th element:
\vspace{-3pt}
\[
\frac{\partial \tau_n[t]}{\partial a_k^n}=
sin\left(\frac{\pi{k}(t-1)}{(T-1)}\right). \tag{14}
\vspace{-3pt}
\]

\textbf{Update}: We use the Adam algorithm~\cite{kingma2014adam} with the step size of $0.01$ to update the learning parameters. The Adam algorithm requires $\frac{\partial \mathcal{D}}{\partial \bm{A}}$ obtained in the backward step and it performs $\mathcal{O}(NK)$ arithmetic operations to update the parameters, where $NK$ is the number of learning parameters.

\textbf{Computational Complexity of TTW -- } Calculating Equation~(12) is the most computationally expensive step in the TTW algorithm. Time and space complexities needed to calculate this step is $\mathcal{O}(KNT)$. We run this step in a loop for $I$ iterations ($I=100$ in our experiments); therefore, time and space complexities of TTW are $\mathcal{O}(IKNT)$ and $\mathcal{O}(KNT)$ respectively. According to our experiments, a good choice for $K$ is a number between $8$ and $16$ for the datasets of the UCR archive.
\vspace{-6pt}
\section{Experiments}
\vspace{-3pt}
\label{sec:experiments}

We use the UCR (University of California, Riverside) time-series classification archive to conduct experiments of this paper~\cite{UCRArchive}. This archive contains $85$ datasets from a wide variety of applications (e.g., medical imaging, geology and astronomy). Each dataset is split into train and test sets and both sets contain class labels up to 60 classes. All time-series in a dataset of UCR have the same length; for different datasets this length varies from 24 to 2,709 samples.

\vspace{2pt}
\textbf{Baseline System}: We implement the GTW algorithm, proposed by Zhou et al.~\cite{zhou2012generalized}, as the baseline method for comparison.
The basis functions and other parameters of the implemented GTW are consistent with Section~5.4 of their paper~\cite{zhou2012generalized}.

\begin{table}
\centering
\caption{Results of the DTW averaging experiment. AVG refers to the conventional sample-by-sample averaging technique. We report two numbers in each cell $(\alpha\text{-}\beta)\%$: $\alpha$ is the percentage of the datasets on which the TTW system is significantly better; $\beta$ is the percentage that the baseline is significantly better. $K$ denotes the number of DST coefficients in Equation~(11). Optimal K refers to the experiment in which we validate $K$ for each dataset.}
\vspace{-6pt}
\label{tab:DTW_avg}
\setlength{\tabcolsep}{3.7pt}
\label{tab:test_ccc_sewa}
\begin{tabular}{c|cccc;{3pt/1pt}c}
TTW      & K = 4 & K = 8 & K = 16 & K = 32 & Optimal K \\
\hline
AVG      & (65-13)\% & (66-15)\% & {(72-19)}\% & (62-26)\% & \textit{\textbf{(82-4)}}\% \\
GTW      & (49-31)\% & {(53-31)}\% & (53-40)\% & (38-55)\% & \textit{\textbf{(57-25)\%}}\\
\end{tabular}
\vspace{-15pt}
\end{table}

\vspace{-6pt}
\subsection{DTW Averaging Experiment}
\vspace{-3pt}
\label{sec:Averaging_Experiment}

The goal of DTW averaging task is to find a time-series that have the minimum DTW distance with the given signals~\cite{niennattrakul2007inaccuracies, petitjean2016faster}. In this section, we compare three time-series averaging techniques (i.e., TTW, GTW and AVG) for the DTW averaging task. AVG is the straightforward sample-by-sample averaging technique. We selected GTW and AVG as the baseline systems because their computational complexity is similar or better than TTW.

For each class of a dataset, we randomly select ten sets of time-series, each set contains ten signals. We estimate the DTW average signal of each set using TTW, GTW and AVG methods. We calculate the DTW distance (Equation~(1)) for each generated average signal. We finally compare our proposed method with baseline systems (i.e., GTW and AVG) in each dataset by applying pair-wise t-test on the DTW distances. For each dataset, we identify if the systems are comparable (two-sided p-value $> 0.05$) or one of them (the one with lower mean DTW distance) is significantly better. 

Table~\ref{tab:DTW_avg} shows the percentage of the datasets on which TTW vs. the other baseline system is significantly better. In this table, we report the results of TTW with different numbers of DST coefficients (i.e., $K$).   According to Table~\ref{tab:DTW_avg}, TTW outperforms both GTW and AVG for all values of $K$ except $32$. GTW approximates the temporal warping by combining a set of basis functions. The results show that the GTW algorithm is not flexible enough to find optimal warping functions in many datasets; however, TTW offers a more flexible time warping that is able to learn complex temporal mappings.

\begin{figure}[t]
\centering
\def\factora{1}
\def\factorb{1}
\begin{subfigure}{.48\linewidth}
  \centering
  \includegraphics[width=\factora\linewidth]{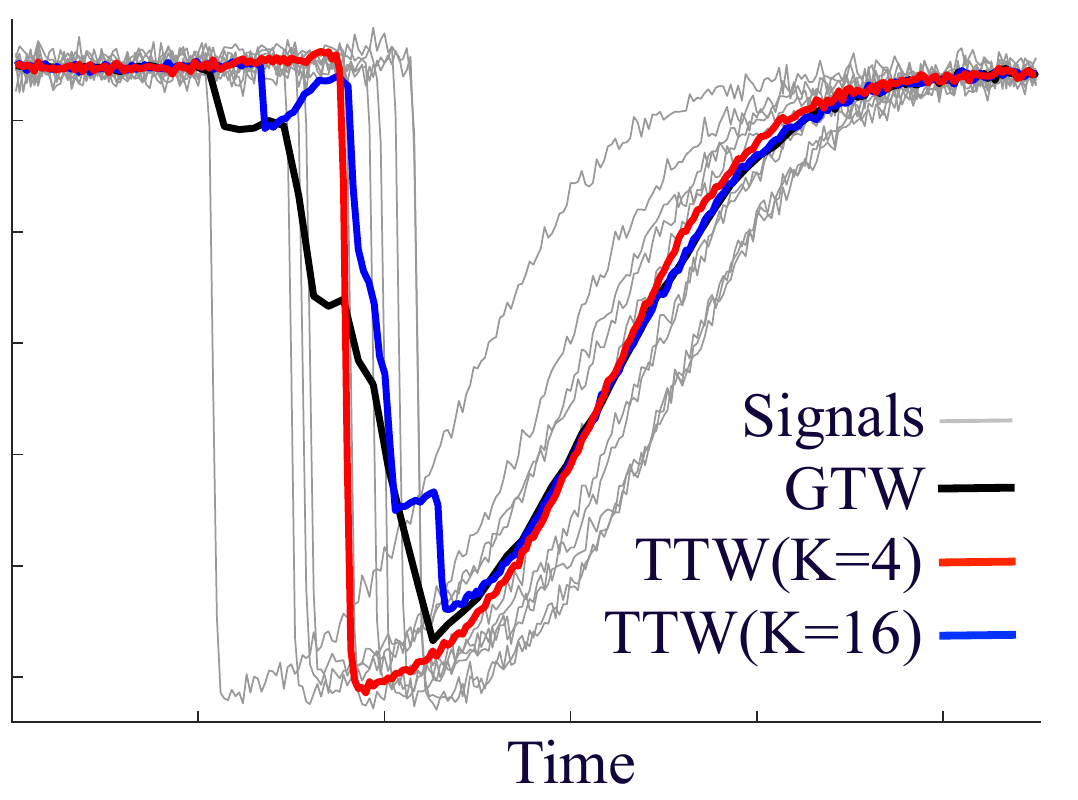}
  \label{fig:sfig1}
\end{subfigure}%
\begin{subfigure}{.52\linewidth}
  \centering
  \includegraphics[width=\factora\linewidth]{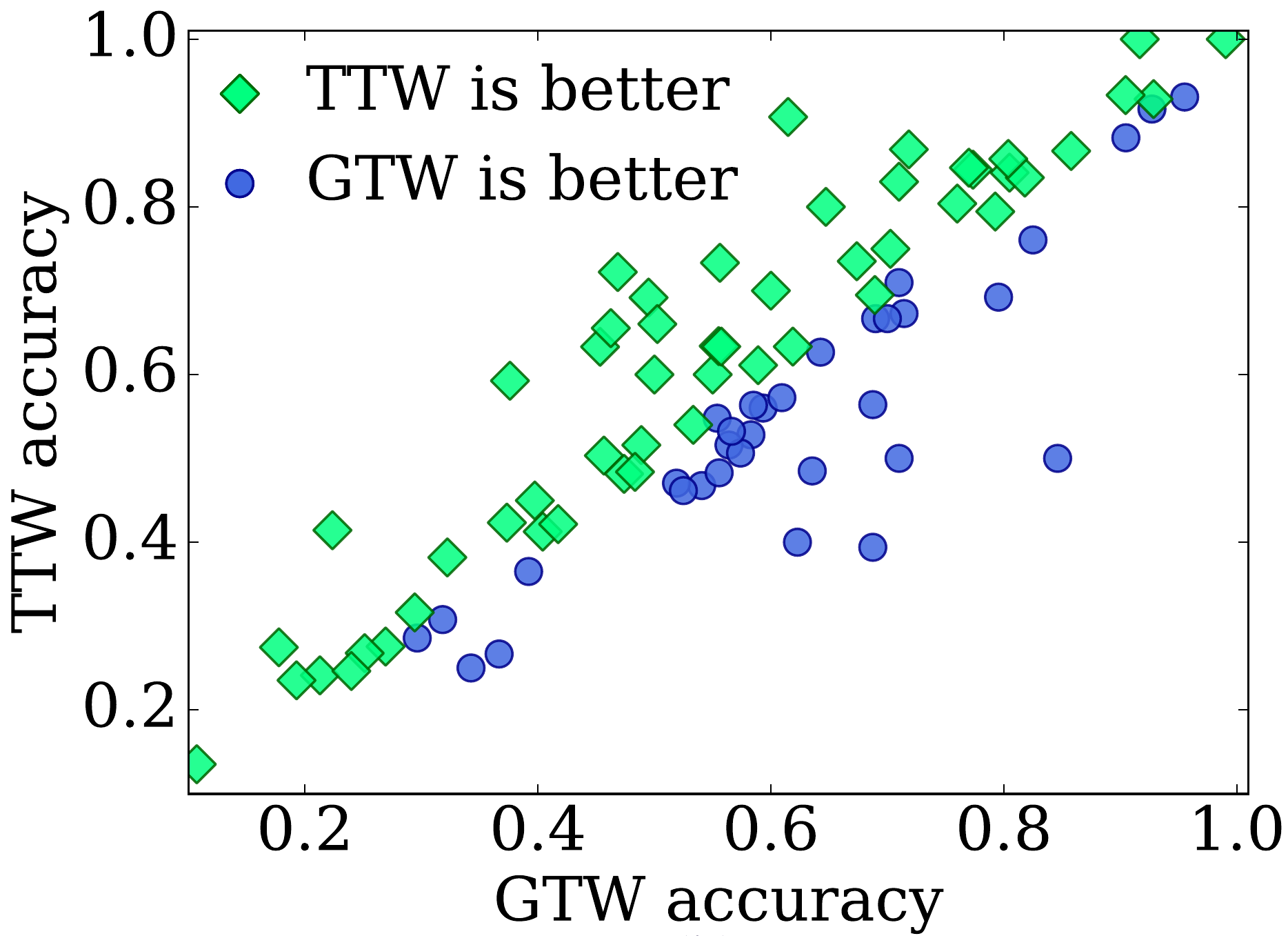}
  \label{fig:sfig2}
\end{subfigure}%
\\\vspace{-0.3cm}\hspace{0.26cm}\textbf{(a)}\hspace{4.1cm}\textbf{(b)}
\vspace{-0.26cm}
\caption{\textbf{(a)} An example of applying DTW averaging techniques on ten randomly selected signals. \textbf{(b)} Results of the nearest centroid classification experiment. Each point reports the result of one dataset; Green rectangular nodes ($52$ out of $85$ datasets) represent the datasets on which TTW outperforms GTW.}
\vspace{-0.7cm}
\label{fig:classification accuracy}
\end{figure}

The performance of the TTW improves by increasing $K$ up to $8$; however, we observe diminishing gains in performance for $K>16$. It is because TTW with a smaller $K$, estimates a smoother warping function that is able to smooth out local optima and provide a better optimization landscape. Figure~\ref{fig:classification accuracy}(a) shows an example of aligning ten time-series using GTW, TTW(K=4) and TTW(K=16). In this example TTW(K=16) converges to a local minimum and cannot find its best solution; however, TTW(K=4) provides a better alignment and considerably outperforms the other methods. 

It is important to run TTW with a reasonable value of $K$: TTW with a small $K$ is not flexible enough to estimate complex warping functions and TTW with a large $K$ is more likely to converge to a local minimum. A straightforward method to define $K$ is to tune it on a subset of the dataset. We use the training sets to tune the value of $K$ and we compare different methods on the test sets. We increase $K$ exponentially from $2^0$ to $2^4$ and select the one that minimizes the DTW distance of the training set of each dataset. We found that by tuning the value of $K$ for each dataset, our system outperforms GTW on $67\%$ of the datasets (significantly for $57\%$ of the datasets). Table~\ref{tab:DTW_avg} (Optimal $K$ column) summarizes the results of this experiment; the results confirm the importance of tuning $K$ for each dataset.

\vspace{-9pt}
\subsection{Classification Experiment}
\vspace{-4pt}

In this section we investigate the performance of the TTW algorithm in a nearest centroid classification application~\cite{li2017distance}. 
We divide the training part of each UCR dataset into two equal parts randomly. We use the first part to find the centroid time-series using TTW and GTW methods. We validate the parameter $K$ on the second part. We select $K$ from five log scale values between $2^0$ and $2^4$. We evaluate our centroids using the UCR test partition. In the test phase, conventional DTW distance assigns each sample to a centroid.

Figure~\ref{fig:classification accuracy}(b) summarizes the results of this experiment. There are $85$ points in this figure; each point compares the classification accuracy of TTW and GTW on a specific dataset. The green rectangular points represent a database for which our proposed algorithm outperforms the GTW algorithm. The results show that the TTW algorithm is able to improve the GTW classification in $61.2\%$ of the datasets. 


\vspace{-9pt}
\section{Conclusion}
\vspace{-4pt}
\label{sec:conclusion}

This paper offers a new solution to the problem of aligning multiple time-series. We first introduce a convolutional sinc kernel that is able to apply non-linear temporal warping functions to time-series. We then use this kernel along with a gradient-based optimization technique to develop a new multiple sequence alignment algorithm: trainable time warping (TTW). The computational complexity of TTW is linear with the number and the length of the input time-series. Our experiments show that TTW provides an effective time alignment; it outperforms GTW (the previous time-warping algorithm that has similar computational complexity) on DTW averaging and nearest centroid classification tasks.

In this paper, we used a sinc filter to approximate the dirac-delta function. Dirac-delta can also be approximated by other functions such as Gaussian. Future work will explore the effect of using different types of kernels.


\vspace{-8pt}
\section{Acknowledgement}
\vspace{-4pt}
This work was partially supported by the National Science Foundation (NSF CAREER 1651740), NIMH R34MH100404, the HC Prechter Bipolar Research Program, and  Richard Tam Foundation. 

\bibliographystyle{IEEEbib}
\bibliography{refs}

\end{document}